\documentclass[letterpaper, 10 pt, conference]{ieeeconf}  

\usepackage{graphicx}

\makeatletter
\let\MYcaption\@makecaption
\makeatother

\usepackage[font=footnotesize]{subcaption}

\makeatletter
\let\@makecaption\MYcaption
\makeatother

\usepackage[hidelinks]{hyperref}
\usepackage{glossaries-extra}
\usepackage{bm}
\usepackage{multirow}
\usepackage{booktabs}
\usepackage{mathtools}

\newcommand{\figref}[1]{Fig.~\ref{#1}}
\newcommand{\secref}[1]{Section~\ref{#1}}
\newcommand{\tabref}[1]{Table~\ref{#1}}

\DeclareMathOperator{\expl}{exp_{Lie}}

\DeclareMathOperator{\diag}{diag}

\DeclareMathOperator{\MPD}{MPD}
\DeclareMathOperator{\median}{median}

\IEEEoverridecommandlockouts                              

\usepackage{amsmath} 
\usepackage{amssymb}  

\title{\LARGE \bf
Hydra: Marker-Free RGB-D Hand-Eye Calibration
}

\author{Martin Huber$^{1}$, Huanyu Tian$^{1}$, Christopher E. Mower$^{2}$, Lucas-Raphael M{\"u}ller$^{3}$,\\S\'{e}bastien Ourselin$^{1}$, Christos Bergeles$^{1,*}$ and Tom Vercauteren$^{1,*}$
\thanks{This work was supported by core funding from the Wellcome/EPSRC~[WT203148/Z/16/Z; NS/A000049/1], the European Union’s Horizon 2020 research and innovation programme under grant agreement No 101016985 (FAROS project), and EPSRC under the UK Government Guarantee Extension (EP/Y024281/1, VITRRO).}
\thanks{$^{1}$Martin Huber, Huanyu Tian, S\'{e}bastien Ourselin, Christos Bergeles, and Tom Vercauteren are with the School of Biomedical Engineering \& Imaging Sciences, King's College London, London, UK.
        {\tt\small martin.huber@kcl.ac.uk}}%
\thanks{$^{2}$Christopher E. Mower is with Noah's Ark Lab, Huawei, London, UK. Work performed while with King's College London.}%
\thanks{$^{3}$Lucas-Raphael M{\"u}ller is an independent contributor.}%
\thanks{$^{*}$Authors contributed equally}%
}

\setabbreviationstyle[acronym]{long-short}
\glssetcategoryattribute{acronym}{nohyperfirst}{true}

\newacronym{dof}{DoF}{degrees of freedom}
\newacronym{dr}{DR}{differentiable rendering}
\newacronym{icp}{ICP}{iterative closest point}
\newacronym{irls}{IRLS}{iteratively reweighted least squares}
\newacronym{mad}{MAD}{median absolute deviation}
\newacronym{mpd}{MPD}{mean pairwise distance}
\newacronym{ptp}{PTP}{point-to-plane}
\newacronym{pnp}{PnP}{perspective-n-point}
\newacronym{tcp}{TCP}{tool center point}
\newacronym{vlm}{VLM}{vision-language model}
\begin{document}

\maketitle
\thispagestyle{empty}
\pagestyle{empty}

\begin{abstract}
This work presents an RGB-D imaging-based approach to marker-free hand-eye calibration using a novel implementation of the \gls{icp} algorithm with a robust \gls{ptp} objective formulated on a Lie algebra. Its applicability is demonstrated through comprehensive experiments using three well known serial manipulators and two RGB-D cameras. With only three randomly chosen robot configurations, our approach achieves approximately $90\%$ successful calibrations, demonstrating $2-3\times$ higher convergence rates to the global optimum compared to both marker-based and marker-free baselines. We also report $2$ orders of magnitude faster convergence time ($0.8\pm0.4\,\text{s}$) for $9$ robot configurations over other marker-free methods. Our method exhibits significantly improved accuracy ($5\,\text{mm}$ in task space) over classical approaches ($7\,\text{mm}$ in task space) whilst being marker-free. The benchmarking dataset and code are open sourced under Apache 2.0 License, and a ROS 2 integration with robot abstraction is provided to facilitate deployment\footnote{\label{fn:source}Software: \url{https://github.com/roboreg}}$^{,}$\footnote{Benchmark dataset: \url{https://drive.google.com/file/d/1WAU9EgYlW-wVcVWL_4bldZYYWfXXX5eM/view?usp=sharing}}.
\end{abstract}

\section{Introduction}
\label{sec:introduction}
Hand-eye calibration -- determining a camera's pose relative to a serial manipulator and thus relating joint space to a Cartesian reference space -- is a core component of robotics and computer vision, enabling autonomous operation and structured interaction with the environment.
Task specification in Cartesian space minimizes redundancy~\cite{action_space} and remains agnostic to the robot’s kinematics~\cite{universal_manipulation_interface}, enhancing efficiency and enabling more generalizable data-driven policy learning. Given the inherent data scarcity in robotics, effective hand-eye calibration remains essential for achieving autonomy.
\begin{figure}[t]
        \centering
        \begin{subfigure}{0.49\columnwidth}
                \centering
                \includegraphics[width=0.9\textwidth]{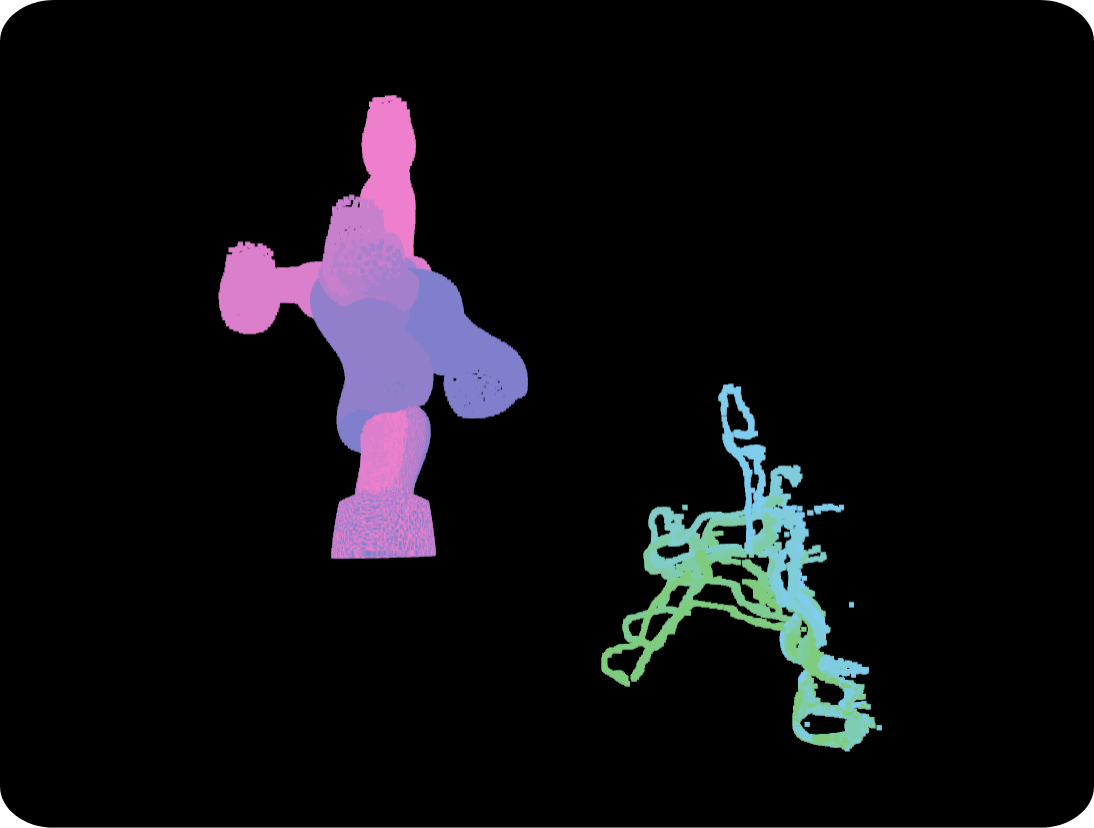}
                \caption{Unaligned robot model.}
                \label{fig:hydra_unregistered}
        \end{subfigure}
        \begin{subfigure}{0.49\columnwidth}
                \centering
                \includegraphics[width=0.9\textwidth]{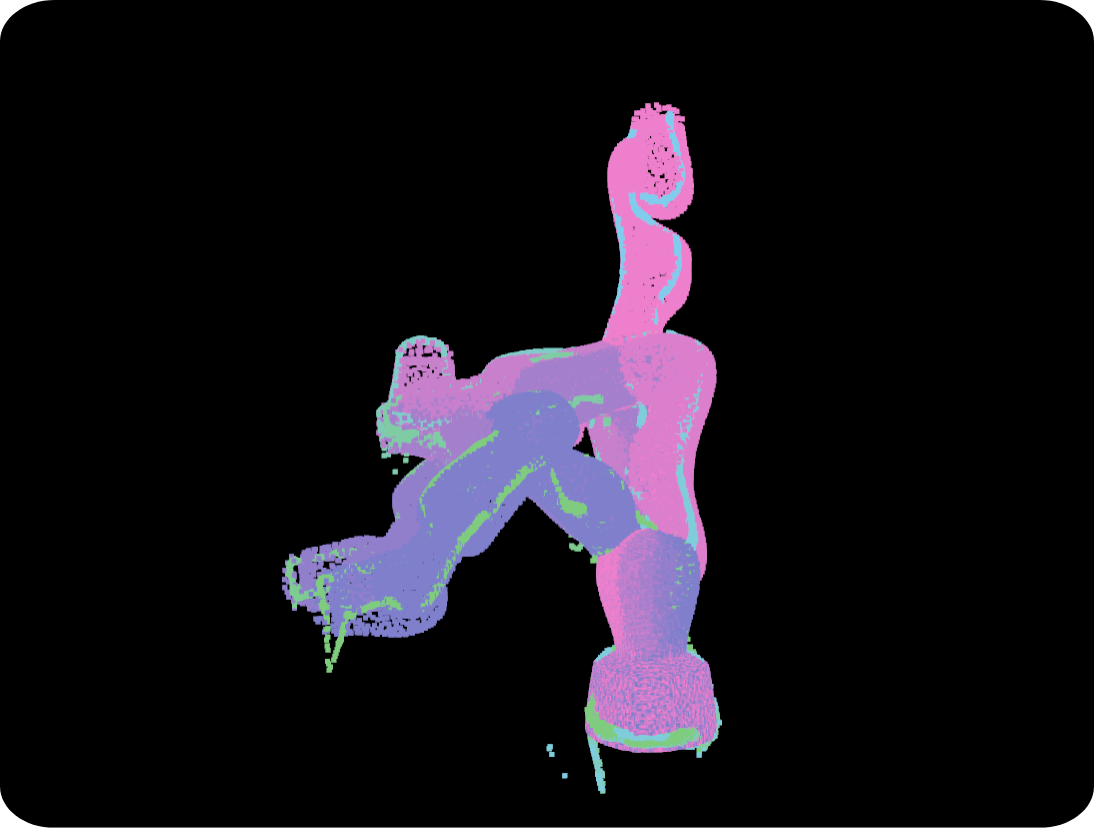}
                \caption{Aligned robot model.}
                \label{fig:hydra_registered}
        \end{subfigure}
        \caption{Hydra takes advantage of multiple robot configurations. It aligns robot mesh vertices (purple) with observed point clouds (turquoise). The combined robot configurations resemble the mythical Hydra.
        }
        \label{fig:hydra_sample}
\end{figure}

As detailed in \secref{sec:experiments.protocol}, classical approaches to hand-eye calibration rely on markers. However, markers 
are subject to strict visibility constraints, and add complexity to the calibration process for users. Consequentially, a significant body of literature aims to alleviate these limitations by using 
the robot 
itself
as a marker surrogate.
In this work, we refer to these approaches as marker-free. Summarizing requirements, the ideal hand-eye calibration should be accurate in task space, marker-free, robust, applicable to any serial manipulator, and expedient. 
\begin{figure*}
        \centering
        \includegraphics[width=0.8\textwidth]{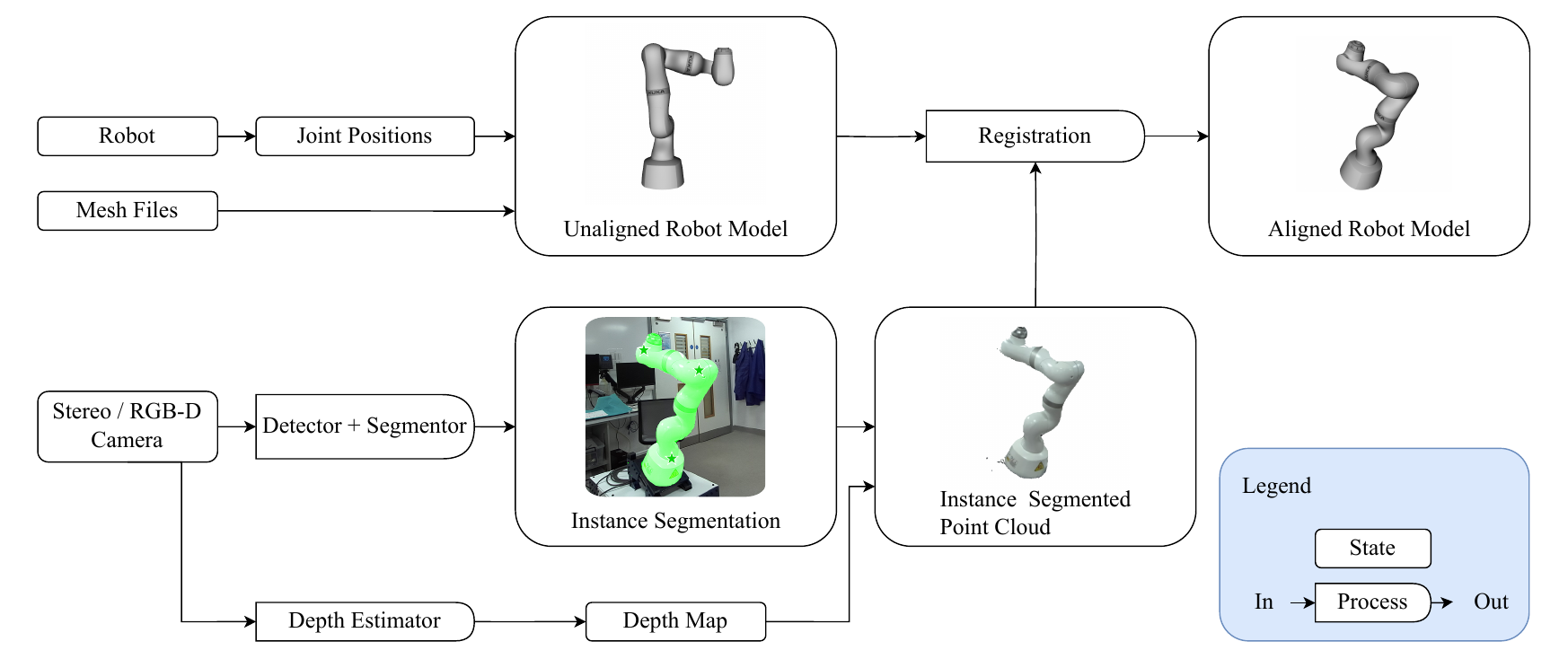}
        \caption{Schematic of the proposed Hydra method. Refers to \secref{sec:methods}.}
        \label{fig:hydra_schematic}
\end{figure*}

In~\cite{nvidiapnp}, it was demonstrated how supervised learning on synthetic data allows for keypoints extraction from the robot, enabling 2D-3D correspondence finding and calibration via \gls{pnp}. This idea was further improved towards self-supervised keypoint identification in CtRNet~\cite{ctrnet}. Both works rely on training dedicated models and applicability to any robot was not demonstrated. Recently, CtRNet-X was introduced in~\cite{ctrnetx} to improve outlier robustness, but it directly inherits the limitations of CtRNet.
Alternative \gls{pnp}-based approaches~\cite{kalib} rely on a known \gls{tcp}, are thus incompatible with variable end-effectors, and remain constrained by visibility requirements. 

Beyond keypoint exploitation, several point cloud registration based approaches exist for marker-free hand-eye calibration~\cite{automatic_hand_eye,arbitrary_object}.
These methods either rely on learning to distinguish the robot point cloud or require additional, albeit arbitrary, known objects within the scene. 
Closest to our work, in that they identify the robot via foundation segmentation models in image space~\cite{ravi2024sam2segmentimages},
are EasyHeC and EasyHeC++~\cite{easyhec,easyhec_pp}. 
However, both rely on \gls{dr} and do not address the question of unknown initialization, leading to significant runtime overhead and a higher risk of convergence to local minima.

To enhance runtime efficiency and robustness, we propose directly registering robot meshes to SAM 2-segmented point clouds, rather than performing \gls{dr} on mesh projections (see \figref{fig:hydra_sample} and \figref{fig:hydra_schematic}).
To this end, this paper contributes:
\begin{itemize}
        \item A novel, sample-efficient and computationally efficient \gls{icp} algorithm, incorporating a robust \glsfirst{ptp} objective formulated on a Lie algebra, with applicability to any serial manipulator.
        \item Comprehensive multi-system evaluation on real hardware, accompanied by an openly available benchmark dataset.
        \item Open-source implementation$^{\ref{fn:source}}$, including ROS 2 integration for seamless deployment.
\end{itemize}


\section{Methods}
\label{sec:methods}
Mathematical notation is introduced in \secref{sec:methods.notation}, followed by a formal derivation of the proposed Hydra method in \secref{sec:methods.hydra}. \secref{sec:methods.point_clouds} explains how we obtain point clouds from depth maps and segmentations. \secref{sec:methods.implementation} provides algorithmic implementation remarks. An overview of the proposed algorithm is given in \figref{fig:hydra_schematic}.
\subsection{Mathematical Notation}
\label{sec:methods.notation}
Throughout the paper, scalars are denoted as lower case Greek / Latin symbols (e.g. $\sigma,s$), vectors as lower case bold Greek / Latin symbols (e.g. $\boldsymbol{\sigma},\mathbf{s}$), and matrices as capital case bold Greek / Latin symbols (e.g. $\boldsymbol{\Sigma},\mathbf{S}$). A vector in homogeneous coordinates is denoted with a tilde (e.g. $\tilde{\boldsymbol{\sigma}}=\begin{bmatrix}\boldsymbol{\sigma}&1\end{bmatrix}^T$). The cross product between two vectors is represented by a skew-symmetric matrix (e.g. $\mathbf{a}\times\mathbf{b}=\mathbf{a}_{\times}\mathbf{b}$).

\subsection{Hydra: Robust Point-to-Plane ICP on Lie Algebra}
\label{sec:methods.hydra}
\subsubsection{Objective}
We seek to align a robot mesh through a rigid body transformation $\boldsymbol{\Theta}\in\text{SE}(3)$ 
with an observed point cloud, thus effectively solving for eye-to-hand calibration (cf. \figref{fig:hydra_sample}).
We frame the optimization as a \gls{ptp} objective.
Given point correspondences $\mathcal{C} = \{\left(\tilde{\mathbf{m}}_i,\tilde{\mathbf{o}}_i\right)|i\in\mathcal{I}\}$ (see \secref{sec:methods.implementation}), we optimize for
\begin{equation}
        \sum_{i\in\mathcal{I}}||\tilde{\mathbf{n}}_i^T\left(\boldsymbol{\Theta}\tilde{\mathbf{o}}_i-\tilde{\mathbf{m}}_i\right)||^2=\sum_{i\in\mathcal{I}}\left(\tilde{\mathbf{n}}_i^T\boldsymbol{\Theta}\tilde{\mathbf{o}}_i-\tilde{\mathbf{n}}_i^T\tilde{\mathbf{m}}_i\right)^2
        \label{eq:ptp}
\end{equation}
where $\tilde{\mathbf{n}}_i$ denote normals on the mesh vertices $\tilde{\mathbf{m}}_i$, and $\tilde{\mathbf{o}}_i$ denote points in the observed point cloud. We aim to minimize \eqref{eq:ptp} using an \gls{icp} approach~\cite{ptp_icp}.

\subsubsection{Iterative Optimization on the SE(3) Lie Group}
To guarantee $\boldsymbol{\Theta}$ be in SE(3), we suggest to solve for it in a Lie group iterative optimization approach~\cite{lie_group_iterative:mahony,lie_group_iterative:vercauteren}. The objective in \eqref{eq:ptp} becomes
\begin{equation}
        \sum_{i\in\mathcal{I}} \left( \tilde{\mathbf{n}}_i^T\boldsymbol{\Theta}_j\expl\left(\delta\boldsymbol{\theta}\right)\tilde{\mathbf{o}}_i-\tilde{\mathbf{n}}_i^T\tilde{\mathbf{m}}_i\right)^2
        \label{eq:ptp_lie}
\end{equation}
where given the current estimate $\boldsymbol{\Theta}_j$, we now seek to find an increment $\delta\boldsymbol{\theta}\in\mathfrak{se}(3)$ at each step $\boldsymbol{\Theta}_{j+1}=\boldsymbol{\Theta}_j\expl\left(\delta\boldsymbol{\theta}\right)$.

\subsubsection{Linearization of the Lie Group Objective}
To solve for the increment $\delta\boldsymbol{\theta}=\begin{bmatrix}\delta\boldsymbol{\omega}&\delta\boldsymbol{\tau}\end{bmatrix}^T$, with rotation $\delta\boldsymbol{\omega}\in\mathbb{R}^3$ and translation $\delta\boldsymbol{\tau}\in\mathbb{R}^3$, in a least squares manner, we next linearize the objective in \eqref{eq:ptp_lie}. The exponential map~\cite{se3_tutorial} $\expl:\mathfrak{se}(3)\rightarrow\text{SE}(3)$ has the closed form solution
\begin{equation}
        \expl\left(\delta\boldsymbol{\theta}\right)=\exp\left(\delta\boldsymbol{\theta}_\dagger\right)=\begin{bmatrix}\expl\left(\delta\boldsymbol{\omega}\right)&\mathbf{L}\left(\delta\boldsymbol{\omega}\right)\delta\boldsymbol{\tau}\\\mathbf{0}&1\end{bmatrix}
        \label{eq:exp_map_se3}
\end{equation}
where
\begin{equation}
        \delta\boldsymbol{\theta}_\dagger=\begin{bmatrix}\delta\boldsymbol{\omega}_\times&\delta\boldsymbol{\tau}\\\mathbf{0}&0\end{bmatrix}
        \label{eq:se3_mat_rep}
\end{equation}
is the matrix representation of $\delta\boldsymbol{\theta}$,
\begin{equation}
        \expl\left(\delta\boldsymbol{\omega}\right)=\mathbf{I}_3+\frac{\sin(\delta\omega)}{\delta\omega}\delta\boldsymbol{\omega}_\times+\frac{1-\cos(\delta\omega)}{\delta\omega^2}\delta\boldsymbol{\omega}_\times^2
        \label{eq:exp_map_so3}
\end{equation}
as per
the Rodrigues' formula with $\delta\omega = ||\delta\boldsymbol{\omega}||$, and
\begin{equation}
        \mathbf{L}(\delta\boldsymbol{\omega})=\mathbf{I}_3+\frac{1-\cos(\delta\omega)}{\delta\omega^2}\delta\boldsymbol{\omega}_\times+\frac{\delta\omega - \sin(\delta\omega)}{\delta\omega^3}\delta\boldsymbol{\omega}_\times^2
        \label{eq:L_omega}
\end{equation}
Finally, linearizing \eqref{eq:exp_map_so3} and \eqref{eq:L_omega} for small $\delta\omega$, turns \eqref{eq:exp_map_se3} into
\begin{equation}
        \expl\left(\delta\boldsymbol{\theta}\right)\approx\mathbf{I}_4 + \delta\boldsymbol{\theta}_\dagger
        \label{eq:exp_map_se3_linearized}
\end{equation}
\subsubsection{Lie Algebra Linearization of the PTP Objective}
The action of the exponential $\expl\left(\delta\boldsymbol{\theta}\right)\tilde{\mathbf{o}}_i$ in \eqref{eq:ptp_lie} can now be linearized through \eqref{eq:exp_map_se3_linearized} as follows
\begin{equation}
        \begin{split}
                \expl&\left(\delta\boldsymbol{\theta}\right)\tilde{\mathbf{o}}_i\approx\tilde{\mathbf{o}}_i+\begin{bmatrix}\delta\boldsymbol{\omega}_\times&\delta\boldsymbol{\tau}\\\mathbf{0}&0\end{bmatrix}\tilde{\mathbf{o}}_i\\
                &=\tilde{\mathbf{o}}_i+\begin{bmatrix}-\mathbf{o}_{i_\times}\delta\boldsymbol{\omega}+\delta\boldsymbol{\tau}\\0\end{bmatrix}\\
                &=\tilde{\mathbf{o}}_i+\begin{bmatrix}-\mathbf{o}_{i_\times}&\mathbf{I}_3\\\mathbf{0}&\mathbf{0}&\end{bmatrix}\delta\boldsymbol{\theta}:=\tilde{\mathbf{o}}_i+\mathbf{D}\left(\mathbf{o}_i\right)\delta\boldsymbol{\theta}
        \end{split}
        \label{eq:action_linearized}
\end{equation}
Plugging \eqref{eq:action_linearized} into \eqref{eq:ptp_lie} finally yields
\begin{equation}
        \sum_{i\in\mathcal{I}}\left(\tilde{\mathbf{n}}_i^T\boldsymbol{\Theta}_j\mathbf{D}\left(\mathbf{o}_i\right)\delta\boldsymbol{\theta}+\tilde{\mathbf{n}}_i^T\boldsymbol{\Theta}_j\tilde{\mathbf{o}}_i-\tilde{\mathbf{n}}_i^T\tilde{\mathbf{m}}_i\right)^2
        \label{eq:ptp_lie_lstsq}
\end{equation}
which represents a standard least squares problem
\begin{equation}
        ||\mathbf{A}\delta\boldsymbol{\theta}-\mathbf{B}||^2
        \label{eq:lstsq}
\end{equation}
where we define $\mathbf{a}_i:=\tilde{\mathbf{n}}_i^T\boldsymbol{\Theta}\mathbf{D}\left(\mathbf{o}_i\right)$, $\mathbf{A}:=\left[\mathbf{a}_0;...;\mathbf{a}_{N-1}\right]$, $b_i:=\tilde{\mathbf{n}}_i^T\left(\tilde{\mathbf{m}}_i-\boldsymbol{\Theta}_j\tilde{\mathbf{o}}_i\right)$, and $\mathbf{B}:=\left[b_0;...;b_{N-1}\right]$.
\subsubsection{Robust Formulation}
To account for potential outliers in the observed point cloud $\tilde{\mathbf{o}_i}$, we treat \eqref{eq:lstsq} under an \gls{irls} approach~\cite{irls}. This turns \eqref{eq:lstsq} into
\begin{equation}
        ||\mathbf{W}\left(\mathbf{A}\delta\boldsymbol{\theta}-\mathbf{B}\right)||^2
        \label{eq:reweighted_lstsq}
\end{equation}
where the weights $\mathbf{W}=\diag\left(w_\kappa\left(b_0\right);...;w_\kappa\left(b_{N-1}\right)\right)$ associated with the Huber loss are given through the current residuals $b_i$
\begin{equation}
        w_\kappa(b_i)=\begin{cases}1&\text{if}\,\,|b_i|\leq\kappa\\\kappa/|b_i|&\text{otherwise}\end{cases}
\end{equation}
The typical choice $\kappa=1.345\sigma$ provides robustness whilst retaining appropriate properties when the errors are Gaussian. We estimate the standard deviation of the noise $\sigma$ using the current residuals via the \gls{mad}
\begin{equation}
        \sigma\approx\frac{\median\left(|\mathbf{B}-\median\left(\mathbf{B}\right)|\right)}{0.6745}.
        \label{eq:mad}
\end{equation}

\subsection{Point Clouds from Depth Maps and Segmentations}
\label{sec:methods.point_clouds}
Taking advantage of multiple robot configurations in the objective function \eqref{eq:ptp_lie}, we extract robot point clouds through instance segmentations applied to depth maps and fuse them in a combined point cloud $\{\tilde{\mathbf{o}}_i\}$ (cf. \figref{fig:hydra_schematic}). 
We initially segment the serial manipulator in RGB space through prompting SAM 2~\cite{ravi2024sam2segmentimages}, making the proposed approach generally applicable to any robot. Critically, we only keep the boundaries of segmentations via erosion. The eroded segmentations are then applied to the respective depth maps for masking only relevant areas (see \figref{fig:hydra_sample}). This procedure significantly reduces the complexity of correspondence finding, which scales quadratically (see \secref{sec:methods.implementation}), and favors the surface sliding mechanism of the \gls{ptp} objective \eqref{eq:ptp}.
Using the camera intrinsics $\mathbf{K}$, we then project depth maps into camera coordinates and obtain $\{\tilde{\mathbf{o}}_i\}$.

\subsection{Algorithmic Implementation Remarks}
\label{sec:methods.implementation}
The proposed Hydra algorithm draws its naming from registering robot meshes fused across multiple configurations 
(cf. \figref{fig:hydra_sample}).
We compute forward kinematics using PyTorch Kinematics~\cite{pytorch_kinematics} for each configuration under measured joint states $\mathbf{q}_c$, and transform individual link meshes accordingly. For an initial estimate $\boldsymbol{\Theta}_0$, we compute mesh and point cloud centroids for each configuration $\mathbf{q}_c$, then align them using the Kabsch-Umeyama algorithm~\cite{kabsch}. 
Hydra then runs an outer and an inner optimization loop. Given the current best estimate $\boldsymbol{\Theta}_j$, closest correspondences $\mathcal{C}$ are sought for exhaustively between meshes $\tilde{\mathbf{m}}_i$ and point clouds $\tilde{\mathbf{o}}_i$ by evaluating the pairwise Euclidean distance, but on a per configuration $\mathbf{q}_c$ basis. 
The inner optimization loop then updates $\boldsymbol{\Theta}_{j+1}=\boldsymbol{\Theta}_j\expl\left(\delta\boldsymbol{\theta}\right)$ using the linear least squares solution of \eqref{eq:reweighted_lstsq}.

\section{Experiments}
\label{sec:experiments}
\begin{figure}[t]
        \centering
        \includegraphics[width=0.4\textwidth]{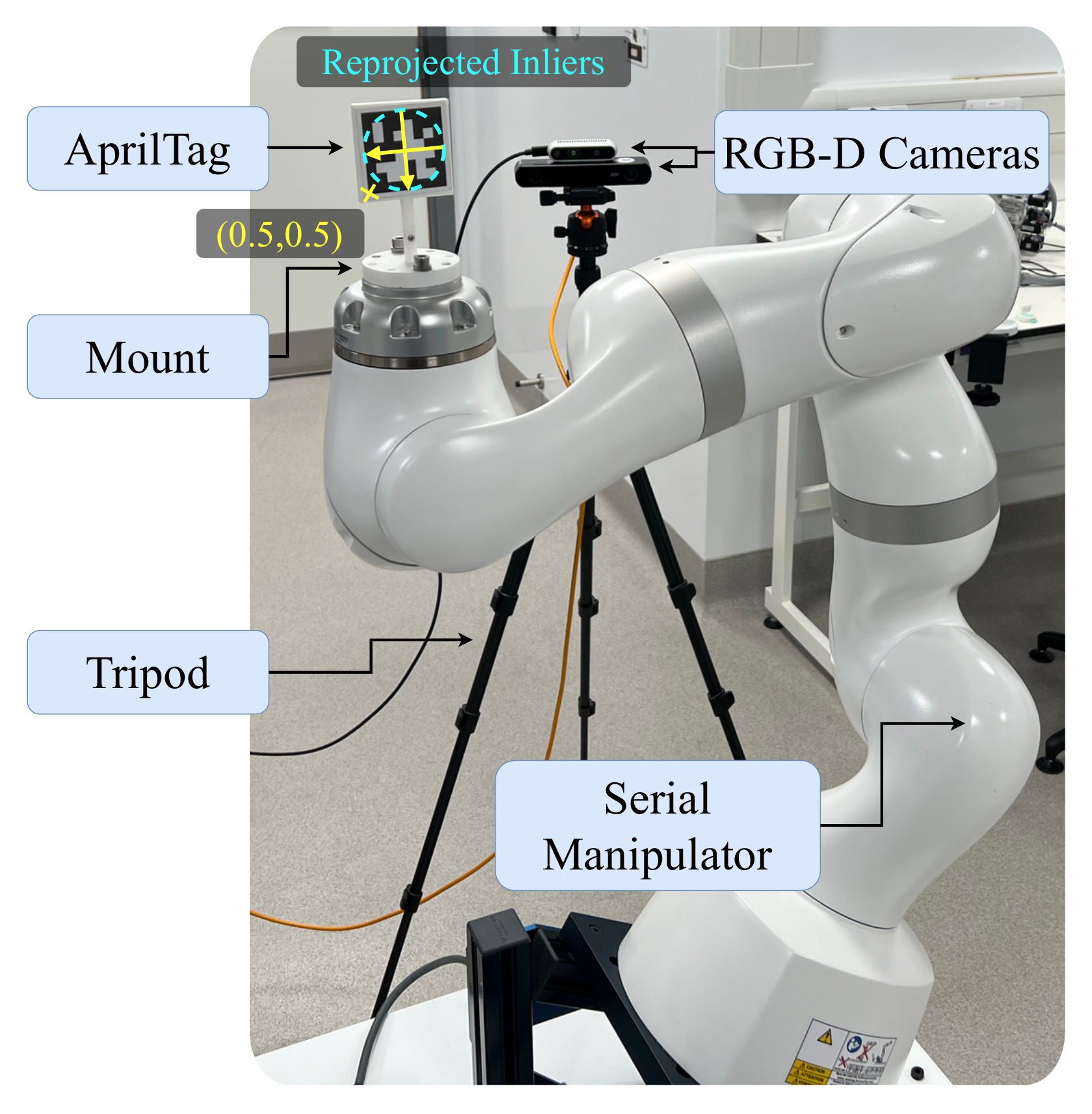}
        \caption{Exemplary experimental setup. A serial manipulator is observed via an RGB-D camera. The yellow axes represent the AprilTag-centric coordinate system for reprojection classification, while the turquoise circle distinguishes inliers from outliers. Refers to \secref{sec:experiments}.}
        \label{fig:experimental_setup}
\end{figure}
We carried out a set of experiments to quantitatively evaluate robustness, repeatability, transferability across different setups, and accuracy in task space (see \secref{sec:experiments.protocol}), for a variety of serial arms. 
We collected a dataset using, independently, two RGB-D cameras, and three serial manipulators (see \figref{fig:experimental_setup}, \secref{sec:experiments.setup}). 
Each serial manipulator was observed from three randomly selected camera poses across 15 randomly chosen joint space configurations, resulting in a total of 270 unique samples.

\subsection{Quantitative Evaluation Protocol}
\label{sec:experiments.protocol}

\subsubsection{Benchmarks} First, we considered classical marker-based approaches, e.g.~\cite{hec_tsai}, that solve for an equation of the form\footnote{where the traditional $\mathbf{A}$ and $\mathbf{B}$ are not to be confused with those in \eqref{eq:reweighted_lstsq}}
$\mathbf{A}\mathbf{X} = \mathbf{X}\mathbf{B}$. 
Robust solvers to this problem were considered~\cite{robust_eye_hand_calibration}, but as no reference implementation was found, we did not include those in our benchmark.
Second, we investigated the classical marker-based eye-to-hand formulation $\mathbf{A}\mathbf{X} = \mathbf{Z}\mathbf{B}$, e.g.~\cite{hec_shah}. We utilized readily available implementations from OpenCV\footnote{\url{https://opencv.org}}. 
Third, we took advantage of our known-by-design AprilTag-to-gripper transformation (cf. \figref{fig:experimental_setup}) and solved for a marker-based \gls{pnp}, where each of the 4 AprilTag edges in each robot configuration was considered a 3D point. 
Fourth, we benchmarked against the marker-free EasyHeC \gls{dr}-based method.

\subsubsection{Quasi Task Space Metric} The robust localization of AprilTag centers $\mathbf{a}^{\prime}_i$ in image space allows for providing precise ground truth. Given the calibrated robot base pose in the camera frame $\prescript{b}{c}{\boldsymbol{\Theta}}$ and the AprilTag 3D position in the robot base frame $\prescript{}{b}{\mathbf{a}_i}$, we can re-project the model 3D AprilTag center into the image plane via $\prescript{}{c}{\tilde{\mathbf{a}}_i}= \prescript{c}{b}{\boldsymbol{\Theta}}\prescript{}{b}{\tilde{\mathbf{a}}}_i$, and $\prescript{}{c}{\tilde{\mathbf{a}}_i}^{\prime}=\frac{1}{z_a}\mathbf{K}\prescript{}{c}{\tilde{\mathbf{a}}_i}$. 
We can then compute the \gls{mpd} in pixels $\MPD\left(\prescript{}{c}{\mathbf{a}^{\prime}_i}, \mathbf{a}^{\prime}_i\right)$. This metric, whilst useful to distinguish errors above and below the camera resolution, depends on the AprilTag scale, which varies with robot and camera configurations. Therefore, we additionally express $\prescript{}{c}{\mathbf{a}^{\prime}_i}$ in coordinates rescaled according to the AprilTag boundaries, centered at $\mathbf{a}^{\prime}_i$ (see  \figref{fig:experimental_setup}). 
Simply computing the average length of $\prescript{}{c}{\mathbf{a}^{\prime}_i}$ in this AprilTag-centric coordinate system and scaling it by the AprilTag size then allow the estimation of task space accuracy. Importantly, our image-based evaluation remains robust even under RGB–depth sensor calibration misalignments, which may occur in structured-light RGB-D cameras. We note however that this approach does not capture  errors along the camera axis.

\subsubsection{Repeatability and Robustness} For repeatability and robustness assessments in task space, we performed Monte Carlo cross validation. For each robot, camera, and camera pose, we randomly selected 5 sets of size $N\in\{3,6,9,12\}$ out of the 15 robot configurations. We then performed the calibration on N samples and tested against the quasi task space metric above on the remaining $15 - N$ samples. We averaged over the sets and propagate errors accordingly.
\subsection{Hardware and Software Setup}
\label{sec:experiments.setup}
\subsubsection{System Properties}
All cameras and serial manipulators were interfaced through a single machine running Ubuntu 22.04 and ROS 2 Humble~\cite{ros2} as the middleware. The machine was equipped with an Intel{\textregistered} Core\textsuperscript{\texttrademark} i7-9750H CPU and an NVIDIA GeForce RTX\textsuperscript{\texttrademark} 2070 with Max-Q Design GPU. The GPU had 8 GB of VRAM but less than 4 GB were used in practice for depth estimation, segmentation, and registration (see \figref{fig:hydra_schematic}). 
\subsubsection{RGB-D Cameras}
The two RGB-D cameras (Intel{\textregistered} RealSense\textsuperscript{\texttrademark} D435 and Stereolabs ZED 2i) were mounted onto a common tripod (see \figref{fig:experimental_setup}). The RealSense\textsuperscript{\texttrademark} D435 camera was rigidly fixated on top of the ZED 2i. 
The RealSense\textsuperscript{\texttrademark} D435 camera was operated at its default resolution of $1280\times720$ pixels through the ROS 2 wrapper\footnote{\url{https://github.com/IntelRealSense/realsense-ros}}. The ZED 2i was operated at its default frame grabbing resolution of $1920\times1080$. However, to reduce bandwidth, its ROS 2 wrapper\footnote{\url{https://github.com/Stereolabs/zed-ros2-wrapper}} downscaled frames to $960\times540$ pixels. The ZED 2i was operated in NEURAL depth mode, which is a proprietary depth estimation algorithm by Stereolabs.
\subsubsection{Serial Manipulators and Calibration Target}
Each of the three serial manipulators had a $5\times5~\text{cm}$ AprilTag~\cite{april_tag} mounted as an end-effector (see \figref{fig:experimental_setup}). We used the apriltag software\footnote{\url{https://github.com/AprilRobotics/apriltag}} for pose estimation. 
First, a KUKA LBR Med 7 R800 was mounted at a 30$^{\circ}$ angle on a cart and controlled through MoveIt~\cite{moveit} configurations of the LBR-Stack~\cite{ros2:lbr_stack}. Second, a UFACTORY xArm 7 \gls{dof} was table-mounted and controlled through MoveIt configurations of the xarm\_ros2 package\footnote{\url{https://github.com/xArm-Developer/xarm_ros2}}. Third, a Mecademic Meca500-R4 was desk-mounted in a laboratory environment and controlled through its Python API\footnote{\url{https://github.com/Mecademic/mecademicpy}} and an admittance controller~\cite{optas}. The admittance was realized via a Bota Systems MiniONE Pro 6-axis force torque sensor, whose cable visually obstructed parts of the Meca500-R4 robot.

\section{Results}
\label{sec:results}
\begin{figure}
        \centering
        \begin{subfigure}{\columnwidth}
                \centering
                \includegraphics[width=0.49\textwidth]{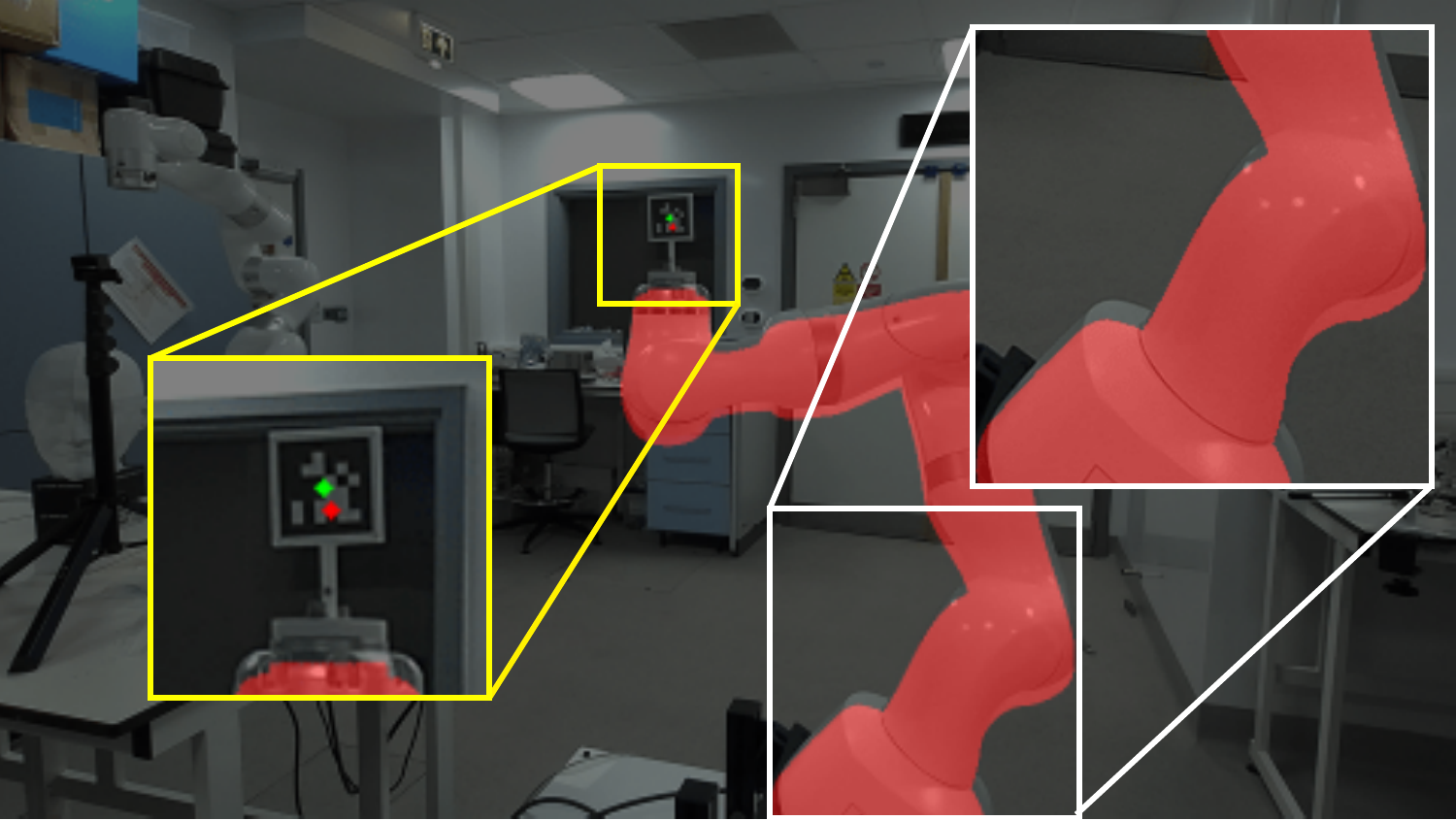}
                \includegraphics[width=0.49\textwidth]{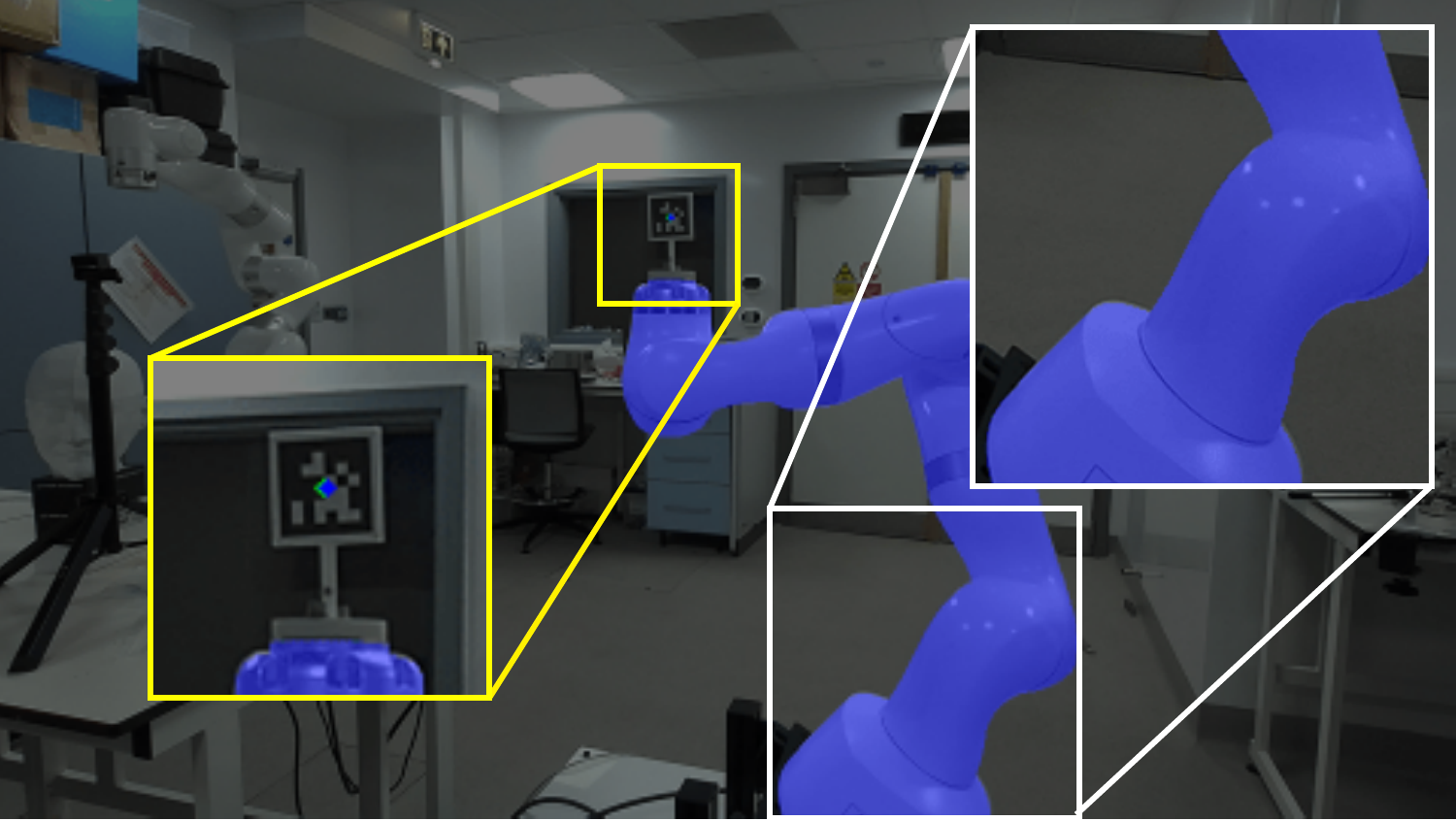}
                \caption{LBR Med 7 and ZED camera, left to right: Shah, Hydra (ours).}
                \label{fig:seg.lbr}
        \end{subfigure}
        \begin{subfigure}{\columnwidth}
                \centering
                \includegraphics[width=0.49\textwidth]{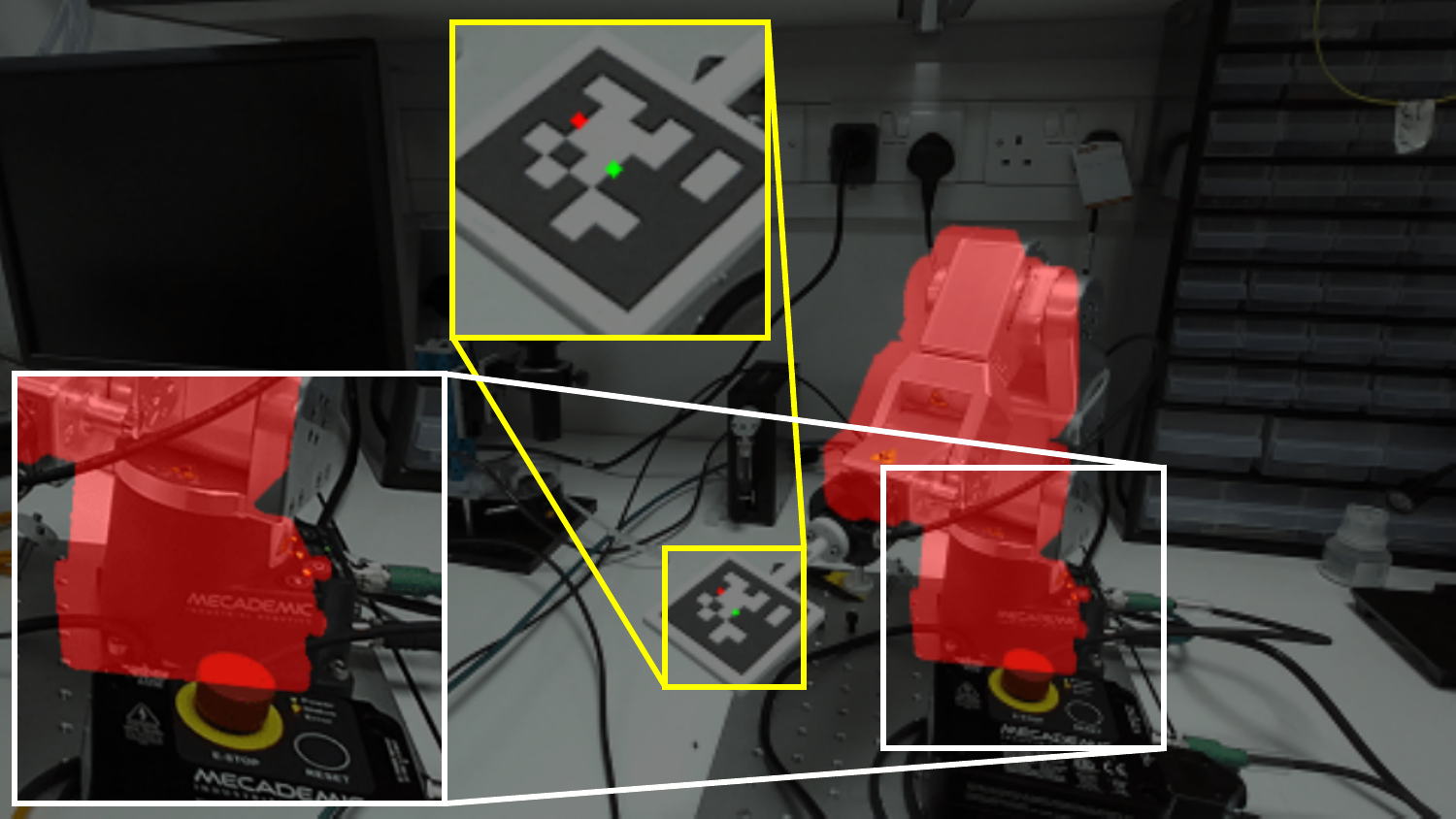}
                \includegraphics[width=0.49\textwidth]{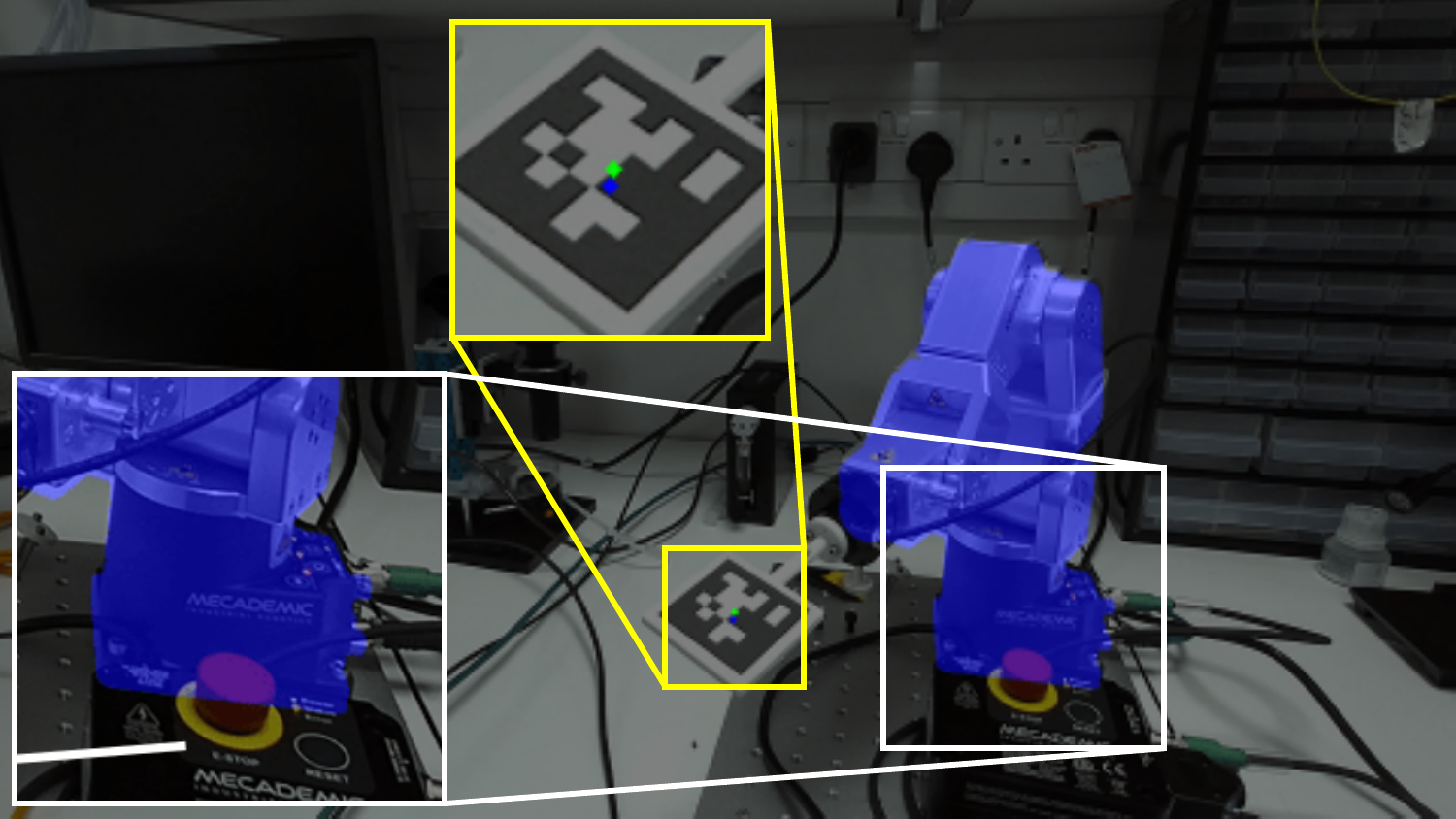}
                \caption{Meca500 and ZED camera, left to right: Shah, Hydra (ours).}
                \label{fig:re_projection_baseline.lbr}
        \end{subfigure}
        \begin{subfigure}{\columnwidth}
                \centering
                \includegraphics[width=0.49\textwidth]{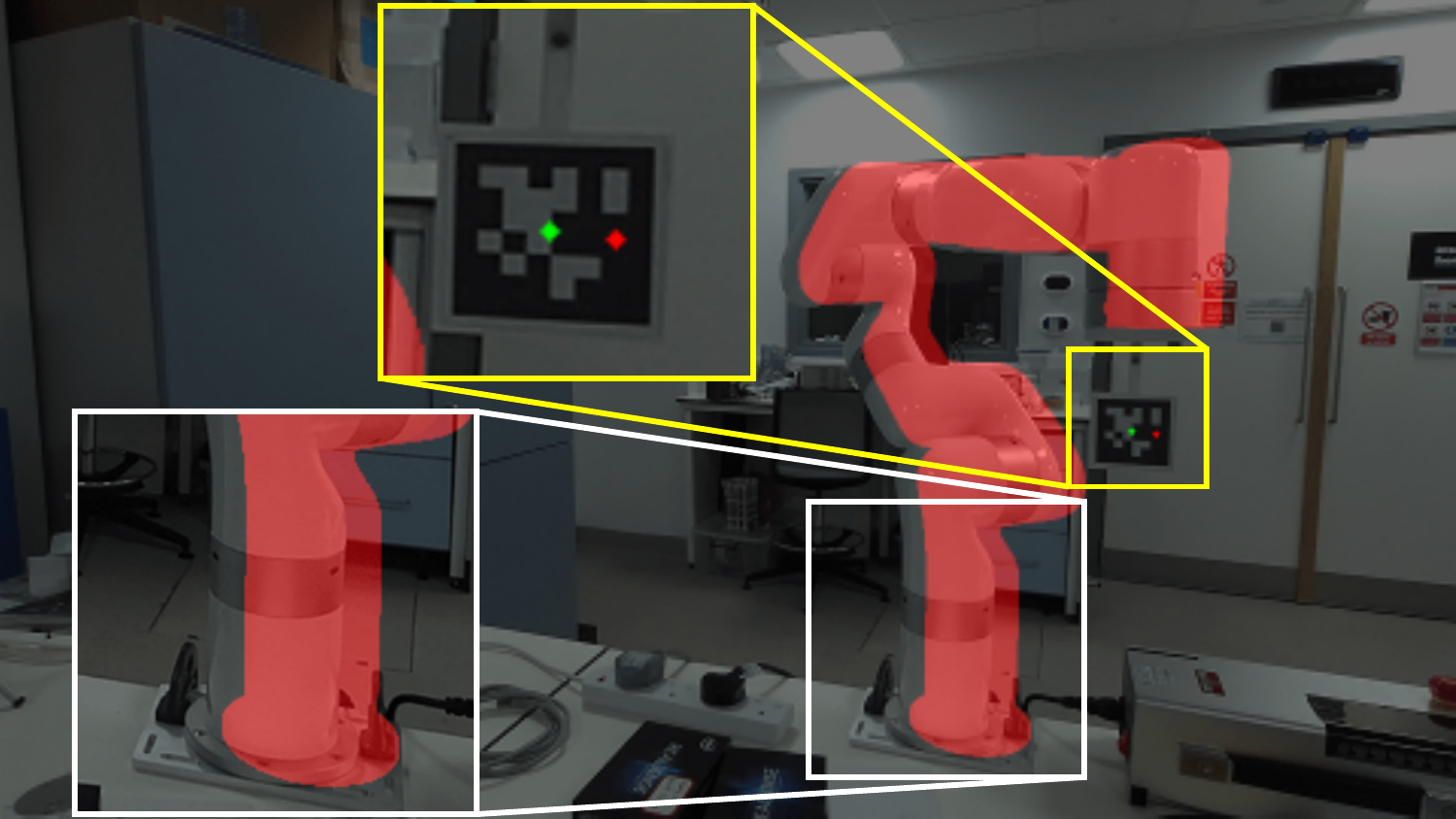}
                \includegraphics[width=0.49\textwidth]{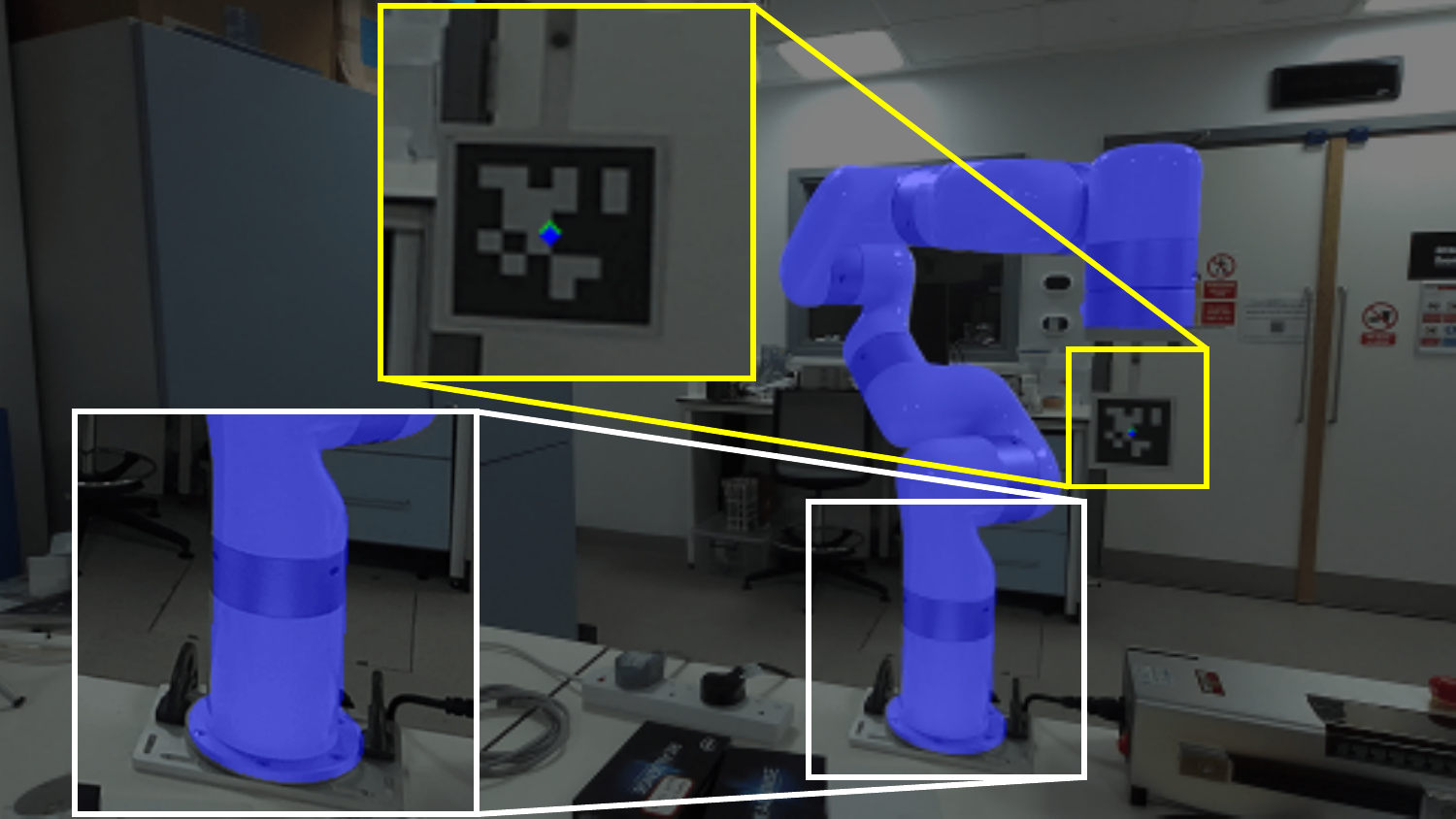}
                \caption{xArm 7 \gls{dof} and ZED camera, left to right: Shah, Hydra (ours).}
                \label{fig:re_projection.lbr}
        \end{subfigure}
        \caption{Meshes rendered using nvdiffrast~\cite{nvdiffrast}. Calibrations for $N=3$ robot configurations. Shown are validation samples, i.e. not among $N$ (refer \secref{sec:experiments.protocol}). Green dots indicate AprilTag centers, red / blue dots indicate reprojected centers. Hydra (ours) is compared to the best classical baseline (Shah~\cite{hec_shah}). Refers to \secref{sec:results.reprojections}.
        }
        \label{fig:re_projection}
\end{figure}

\subsection{Qualitative Reprojections}
\label{sec:results.reprojections}
Qualitative results for $N=3$ robot configurations during calibration are shown in \figref{fig:re_projection}. 
The reprojected AprilTag centers aligned significantly better with the detected ones for Hydra when compared to the classical $\mathbf{A}\mathbf{X} = \mathbf{Z}\mathbf{B}$ baseline of Shah~\cite{hec_shah}. 
One may further observe a much improved alignment of the reprojected meshes with the actual robots. We chose not to display the \gls{pnp} approach, since it relied on known end-effector transformations, which is typically not available in practice. 
It can be seen that Hydra performs slightly worse for the Meca500 robot (also see \figref{fig:cross_validation}). Empirically, we found, as expected, that a relatively larger marker leads to an improved robustness of the marker-based approaches. We further observed that the admittance control with large orientational stiffness limited robot configurations, leading to slightly worse Hydra performance (see \secref{sec:experiments.setup}). Finally, we found that the mesh for the Meca500 robot was relatively sparse, which led to occasional local minima convergence of our method.

\subsection{Quantitative Monte Carlo Cross Validation}
\label{sec:results.cross_validation}

Before presenting the results, we highlight that the available EasyHeC implementation was integrated with the xArm 7 \gls{dof} Python SDK for online calibration, challenging retrospective execution on the collected data (see \secref{sec:experiments}). To allow for retrospective usage of EasyHeC and following their guidelines, we replaced the PointRend segmentation with SAM 2 and substituted the PVNet initialization with the most effective marker-based approach. We denote the adapted version as EasyHeC*. Furthermore, due to the lack of robot abstraction in the reference implementation, the evaluation of EasyHeC was restricted to the xArm 7 \gls{dof}.

\begin{table}[tb]
        \caption{Task space errors (top), raw errors (middle), success rates (bottom). Errors only for successful calibrations (see \figref{fig:experimental_setup}). Monte Carlo cross validation for $N=9$ robot configurations. Refers to \secref{sec:results.cross_validation}.}
        \label{tab:reprojection}
        \resizebox{\linewidth}{!}{
                \begin{tabular}{ll|lllll}
                        \hline
                        Robot                       & Camera    & \begin{tabular}[c]{@{}l@{}}Tsai\\ {[}mm{]}\end{tabular}    & \begin{tabular}[c]{@{}l@{}}Shah\\ {[}mm{]}\end{tabular}    & \begin{tabular}[c]{@{}l@{}}PnP\\ {[}mm{]}\end{tabular}    & \begin{tabular}[c]{@{}l@{}}EasyHeC*\\ {[}mm{]}\end{tabular}    & \begin{tabular}[c]{@{}l@{}}Hydra (ours)\\ {[}mm{]}\end{tabular}    \\ \hline
                        \multirow{2}{*}{LBR Med 7}  & RealSense & $13.4\pm5.5$                                               & $9.3\pm3.9$                                                & $\mathbf{1.8\pm0.9}$                                      & N/A                                                            & $\mathbf{5.3\pm1.8}$                                               \\
                                                    & ZED       & $11.4\pm4.5$                                               & $8.7\pm3.2$                                                & $\mathbf{1.5\pm0.8}$                                      & N/A                                                            & $\mathbf{5.4\pm1.7}$                                               \\ \hline
                        \multirow{2}{*}{Meca500}    & RealSense & $8.4\pm3.0$                                                & $6.5\pm2.0$                                                & $\mathbf{0.6\pm0.3}$                                      & N/A                                                            & $\mathbf{4.2\pm1.5}$                                               \\
                                                    & ZED       & $5.4\pm1.7$                                                & $\mathbf{4.3\pm1.5}$                                       & $\mathbf{0.7\pm0.6}$                                      & N/A                                                            & $6.7\pm2.7$                                                        \\ \hline
                        \multirow{2}{*}{xArm 7 DoF} & RealSense & $10.7\pm4.9$                                               & $6.1\pm2.7$                                                & $\mathbf{3.9\pm2.0}$                                      & $4.9\pm2.6$                                                    & $\mathbf{4.9\pm2.5}$                                               \\
                                                    & ZED       & $6.2\pm3.5$                                                & $4.8\pm2.3$                                                & $\mathbf{3.3\pm1.6}$                                      & $7.5\pm3.6$                                                    & $\mathbf{4.1\pm1.9}$                                               \\ \hline
                        Robot                       & Camera    & \begin{tabular}[c]{@{}l@{}}Tsai\\ {[}pixel{]}\end{tabular} & \begin{tabular}[c]{@{}l@{}}Shah\\ {[}pixel{]}\end{tabular} & \begin{tabular}[c]{@{}l@{}}PnP\\ {[}pixel{]}\end{tabular} & \begin{tabular}[c]{@{}l@{}}EasyHeC*\\ {[}pixel{]}\end{tabular} & \begin{tabular}[c]{@{}l@{}}Hydra (ours)\\ {[}pixel{]}\end{tabular} \\ \hline
                        \multirow{2}{*}{LBR Med 7}  & RealSense & $11.9\pm5.8$                                               & $8.7\pm4.4$                                                & $1.5\pm0.7$                                               & N/A                                                            & $4.2\pm1.3$                                                        \\
                                                    & ZED       & $6.4\pm2.3$                                                & $5.0\pm2.0$                                                & $0.8\pm0.4$                                               & N/A                                                            & $2.8\pm1.0$                                                        \\ \hline
                        \multirow{2}{*}{Meca500}    & RealSense & $13.0\pm4.7$                                               & $10.5\pm2.9$                                               & $1.0\pm0.6$                                               & N/A                                                            & $6.9\pm2.5$                                                        \\
                                                    & ZED       & $4.6\pm1.4$                                                & $3.7\pm1.3$                                                & $0.6\pm0.5$                                               & N/A                                                            & $5.8\pm2.6$                                                        \\ \hline
                        \multirow{2}{*}{xArm 7 DoF} & RealSense & $13.8\pm5.3$                                               & $7.6\pm3.1$                                                & $4.6\pm1.9$                                               & $6.0\pm2.7$                                                    & $6.0\pm2.9$                                                        \\
                                                    & ZED       & $4.8\pm2.6$                                                & $3.9\pm2.0$                                                & $2.5\pm1.1$                                               & $6.0\pm2.7$                                                    & $3.4\pm1.7$                                                        \\ \hline
                        Robot                       & Camera    & \begin{tabular}[c]{@{}l@{}}Tsai\\ {[}a.u.{]}\end{tabular}  & \begin{tabular}[c]{@{}l@{}}Shah\\ {[}a.u.{]}\end{tabular}  & \begin{tabular}[c]{@{}l@{}}PnP\\ {[}a.u.{]}\end{tabular}  & \begin{tabular}[c]{@{}l@{}}EasyHeC*\\ {[}a.u.{]}\end{tabular}  & \begin{tabular}[c]{@{}l@{}}Hydra (ours)\\ {[}a.u.{]}\end{tabular}  \\ \hline
                        \multirow{2}{*}{LBR Med 7}  & RealSense & $0.6$                                                      & $0.8$                                                      & $\mathbf{1.0}$                                            & N/A                                                            & $\mathbf{1.0}$                                                     \\
                                                    & ZED       & $0.47$                                                     & $0.73$                                                     & $\mathbf{1.0}$                                            & N/A                                                            & $\mathbf{1.0}$                                                     \\ \hline
                        \multirow{2}{*}{Meca500}    & RealSense & $\mathbf{1.0}$                                             & $\mathbf{1.0}$                                             & $\mathbf{1.0}$                                            & N/A                                                            & $0.93$                                                             \\
                                                    & ZED       & $\mathbf{1.0}$                                             & $\mathbf{1.0}$                                             & $\mathbf{1.0}$                                            & N/A                                                            & $0.87$                                                             \\ \hline
                        \multirow{2}{*}{xArm 7 DoF} & RealSense & $\mathbf{1.0}$                                             & $\mathbf{1.0}$                                             & $\mathbf{1.0}$                                            & $0.33$                                                         & $\mathbf{1.0}$                                                     \\
                                                    & ZED       & $\mathbf{1.0}$                                             & $\mathbf{1.0}$                                             & $\mathbf{1.0}$                                            & $0.33$                                                         & $\mathbf{1.0}$                                                     \\ \hline
                \end{tabular}
        }
\end{table}

\subsubsection{Robustness, Repeatability, and Computational Efficiency} Detailed results for the Monte Carlo cross validation with $N=9$ robot configurations and $6$ validation samples are listed in \tabref{tab:reprojection}. It can be seen that we operate above the camera resolution (see \tabref{tab:reprojection} center). Despite its lower effective resolution (see \secref{sec:experiments.setup}), the ZED camera scores lower AprilTag center distances than the RealSense camera for marker-based approaches, possibly due to wear of the RealSense camera or more effective autotuning in the ZED software.
Notably, Hydra performs equally well for both cameras, except in the case of the Meca500 robot, particularly when observed with the ZED camera. This is attributed to the Meca500’s sparse meshes (see Section \ref{sec:results.reprojections}) and the lower effective resolution of the ZED camera (see Section \ref{sec:experiments.setup}). Furthermore, it can be seen that Hydra scores lower AprilTag center distances with smaller variations across all baselines, except for the less practical \gls{pnp} approach. In terms of repeatability, \tabref{tab:reprojection} shows that Hydra performs at $100\%$ success rate when the robot target is observed in sufficiently varying configurations. 
Surprisingly, EasyHeC*'s success rate with Shah initialization drops from $100\%$ to $33\%$ for both cameras.
While we deviated from EasyHeC's protocol, we also observe this qualitatively as divergence during minimization (also found in~\cite{kalib}), possibly indicating an ill-parametrized formulation in the published implementation.
To gauge compute time, we measure runtime for each method on the xArm 7 \gls{dof} data. EasyHeC runs for $132.7\pm17.4\,\text{s}$, Hydra for $0.8\pm0.4\,\text{s}$ (which falls well in line with~\cite{automatic_hand_eye}), Hydra without mask erosion for $6.3\pm\,0.5\text{s}$ (\secref{sec:methods.point_clouds}), and Shah for $0.3\pm0.1\,\text{ms}$, where all other marker-based approaches fall under the same order of magnitude. 

\begin{figure}
        \centering
        \includegraphics[width=\linewidth]{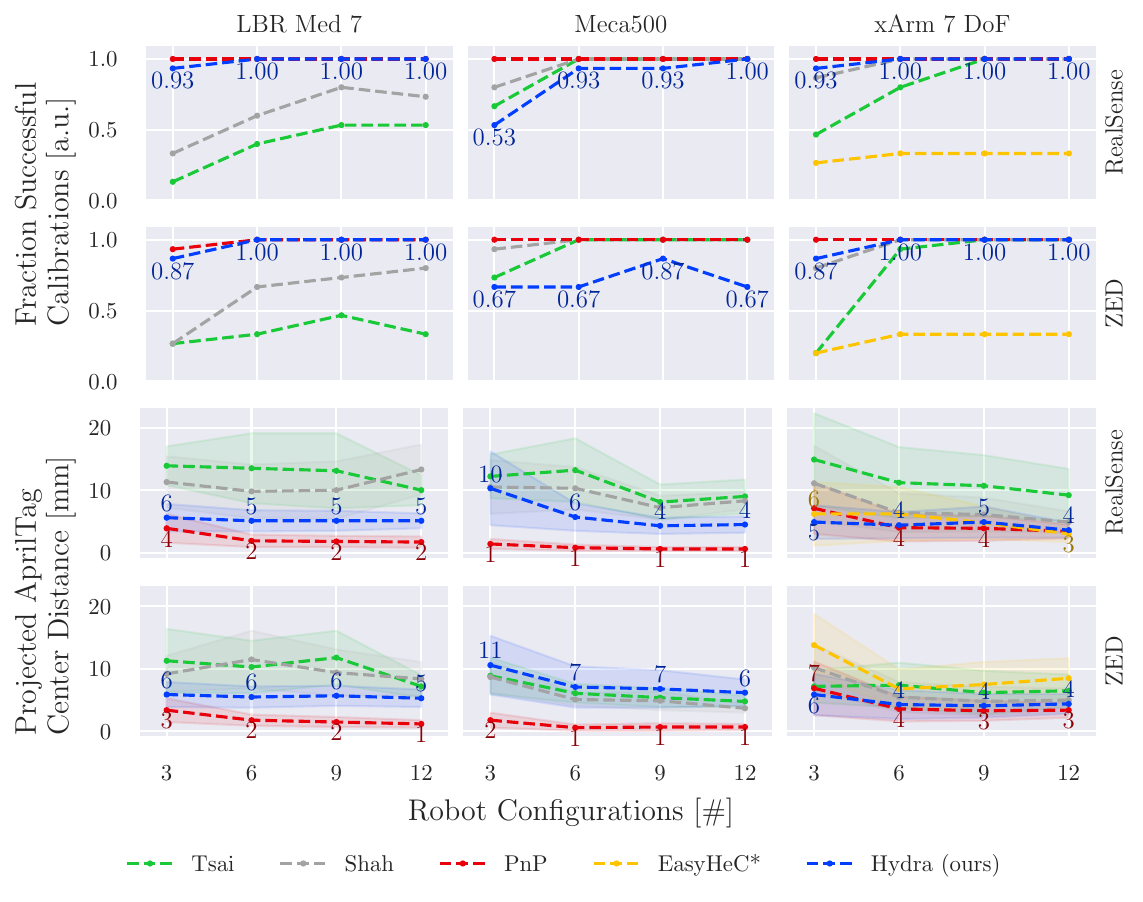}
        \caption{Monte Carlo cross validations for $N\in\{3,6,9,12\}$ robot configurations. Results are displayed by camera and robot (see \secref{sec:experiments.setup}). Fraction of successful calibrations (top two rows, see \figref{fig:experimental_setup}). Only successful calibrations are considered for the reprojection errors (bottom two rows), favoring the less successful methods. Highlighted are the errors for Hydra (ours), and the next best / best approach. Refers to \secref{sec:results.cross_validation}.}
        \label{fig:cross_validation}
\end{figure}

\subsubsection{Robustness and Repeatability with Sample Size} The Monte Carlo cross validation results split by robot and camera across all robot configurations $N\in\{3,6,9,12\}$ are shown in \figref{fig:cross_validation}. Within standard deviation, there is little improvement in accuracy for either method with number of robot configurations. One can, however, notice an improvement in success rates for the marker-based approaches with the number of robot configurations. Except for the Meca500 robot, Hydra saturates its success rate at 6 observed robot configurations. It can further be seen that Hydra performs significantly better for the LBR Med 7 and the Meca500 robot than the practical baselines Shah~\cite{hec_shah} and Tsai~\cite{hec_tsai}. Convergence to local minima, as was explained in \secref{sec:results.reprojections}, decreases Hydra's observed mean for the Meca500 robot. The drop in repeatability for EasyHeC* that was found previously persists across sample sizes. It can further be seen that the most accurate \gls{pnp} approach struggles for the xArm 7 \gls{dof}. This may indicate structural inefficiencies, which is quite common for lightweight manipulators. These structural inefficiencies may induce a lower bound for the investigated methods on the xArm 7 \gls{dof}.

\section{Discussion and Conclusions}
\label{sec:discussion_and_conclusions}
In conclusion, this work demonstrates for the first time that image-space segmentation using a foundation model, similar to EasyHeC, provides a viable abstraction for solving hand-eye calibration across different serial manipulators and cameras. In contrast to EasyHeC, we perform registration in Cartesian space using a novel objective for robust \gls{ptp} \gls{icp} on a Lie algebra. This approach is simpler than \gls{dr} and relies solely on matrix inversion, as shown in \eqref{eq:reweighted_lstsq}, for the linearized objective in \eqref{eq:ptp_lie}. Furthermore, the proposed approach achieves a convergence rate that is two orders of magnitude faster than EasyHeC, with a threefold increase in success rate (see \secref{sec:results.cross_validation} and Fig. \ref{fig:cross_validation}). Additionally, it operates in a truly marker-free manner, eliminating the need for marker-based initialization. Since EasyHeC is only available for the xArm 7 \gls{dof}, and due to observed structural inefficiencies, no final comparison regarding accuracy can be drawn. However, EasyHeC++ reports an error of $3\,\text{mm}$, which is consistent with our reported accuracy (see \figref{fig:cross_validation}). Further, we observe significant improvements over Shah~\cite{hec_shah} and Tsai~\cite{hec_tsai} (see \tabref{tab:reprojection}, \figref{fig:cross_validation}).

A limitation of the proposed Hydra method is its reliance on RGB-D cameras and the SAM 2 segmentation, which requires some, albeit limited, user interaction (see \figref{fig:hydra_schematic} and \secref{sec:methods.implementation}). 
Currently, no effort has been made to adapt the optimization for eye-in-hand setups, and all investigated setups are static, requiring recalibration after displacement. However, it was found that our method exhibits strong accuracy with only three robot configurations (see \figref{fig:re_projection}).

Despite its limitations, Hydra represents a significant contribution to marker-free hand-eye calibration. Future research should focus on dynamic setups and further minimizing the residual error to match the accuracy of the \gls{pnp} method. The open-source nature of the benchmarking dataset and Hydra facilitates advancements in this domain. The mature Python packaging and integration with ROS 2 enable seamless deployment across various robotic systems.







\addtolength{\textheight}{-12cm}   

\bibliographystyle{IEEEtran}
\bibliography{IEEEabrv,references}

\end{document}